\documentclass[akbc,twoside,11pt]{article}
\usepackage{akbc}
\usepackage{siunitx}
\usepackage{color}
\usepackage{algorithm}
\usepackage{algorithmic}
\usepackage{subcaption}
 \usepackage{graphicx}
 \finalcopy

\ShortHeadings{Answering Structured Queries Directly over Text}{Groth,Scerri,Daniel,Allen}

\begin{document}

\title{End-to-End Learning for Answering Structured Queries Directly over Text}

\author{\name Paul Groth \email p.groth@uva.nl \\ University of Amsterdam 
	\AND
       \name Antony Scerri \email a.scerri@elsevier.com \\
       \name Ron Daniel, Jr. \email r.danel@elsevier.com  \\
       \name Bradley P. Allen \email b.allen@elsevier.com  \\
       \addr Elsevier Labs \\
      }


\maketitle

\begin{abstract}
Structured queries expressed in languages (such as SQL, SPARQL, or XQuery) offer a convenient and explicit way for users to express their information needs for a number of tasks. In this work, we present an approach to answer these directly over text data without storing results in a database. We specifically look at the case of knowledge bases where queries are over entities and the relations between them. Our approach combines distributed query answering (e.g. Triple Pattern Fragments) with models built for  extractive question answering. Importantly, by applying distributed querying answering we are able to simplify the model learning problem. We train models for a large portion (572) of the relations within Wikidata and achieve an average 0.70 F1 measure across all models. We also present a systematic method to construct the necessary training data for this task from knowledge graphs and describe a prototype implementation.
\end{abstract}

\section{Introduction}
\label{Introduction}
Database query languages (e.g. SQL, SPARQL, XQuery) offer a convenient and explicit way for users to express their information needs for a number of tasks including populating a dataframe for statistical analysis, selecting data for display on a website, defining an aggregation of two datasets, or generating reports.

However, much of the information that a user might wish to access using a structured query may not be available in a  database and instead available only in an unstructured form (e.g. text documents). To overcome this gap, the area of \emph{information extraction} (IE) specifically investigates the creation of structured data from unstructured content~\cite{martinez2018information}. Typically, IE systems are organized as pipelines taking in documents and generating various forms of structured data from it. This includes the extraction of relations, the recognition of entities, and even the complete construction of databases. The goal then of IE is not to answer queries directly but first to generate a database that queries can be subsequently executed over.

In the mid-2000s, with the rise of large scale web text, the notion of combining information extraction techniques with relational database management systems emerged~\cite{cafarella2007structured,jain2007sql} resulting in what are termed \emph{text databases}. Systems like Deep Dive~\cite{shin2015incremental} InstaRead~\cite{hoffmann2015extreme}, or Indrex~\cite{kilias2015indrex}, use database optimizations within tasks such as query planning to help decide when to perform extractions. While, in some cases, extraction of data can be performed at runtime, data is still extracted to an intermediate database before the query is answered. Thus, all these approaches still require the existence of a structured database to answer the query.

In this paper, we present an approach that \textbf{eliminates the need to have an intermediate database in order to answer structured database queries over text}. This is essentially the same as treating the text itself as the store of structured data. Using text as the database has a number of potential benefits, including being able to run structured queries over new text without the need for a-priori extraction; removing the need to maintain two stores for the same information (i.e. a database and a search index); eliminating synchronization issues; and reducing the need for up-front schema modeling. \cite{alagiannis2012nodb} provides additional rationale for not pre-indexing ``raw data'', although they focus on structured data in the form of CSV files. 

Our approach builds upon three foundations:
\begin{enumerate}
\item the existence of large scale publicly available knowledge bases (Wikidata) derived from text data (Wikipedia);
\item recent advances in end-to-end learning for extractive question answering (e.g. \cite{seo2016bidirectional});
\item the availability of layered query processing engines designed for distributed data (e.g. SPARQL query processes that work over Triple Pattern Fragment \cite{tpfjournal2016} servers).
\end{enumerate}

A high-level summary of our approach is as follows. We use a publicly-available knowledge base to construct a parallel corpus consisting of tuples each which is made up of a structured slot filling query, the expected answer drawn from the knowledge base, and a corresponding text document in which we know the answer is contained. Using this corpus, we train neural models that learn to answer the given structured query given a text document. This is done on a per relation basis. These models are trained end-to-end with no specific tuning for each query. These models are integrated into a system that answers queries expressed in a graph query language directly over text with no relational or graph database intermediary.

The contributions of this paper are:
\begin{itemize}
\item a method for generating training data for the task of answering structured queries over text;
\item models that can answer slot filling queries for over 570 relations with no relation or type specific tuning. These models obtain on average a 0.70 F1 measure for query answering. 
\item a prototype system that answers structured queries using triple pattern fragments over a large corpus of text (Wikipedia).
\end{itemize}

The rest of this paper is organized as follows. We begin with an overview of the approach. This is followed by a description of our method for creating training data. Subsequently, we describe the model training and discuss the experimental results. After which, we present our prototype system.  We end the paper with a discussion of related and future work.

\section{Overview}

\begin{figure}
\centering
\includegraphics[width=.7\textwidth]{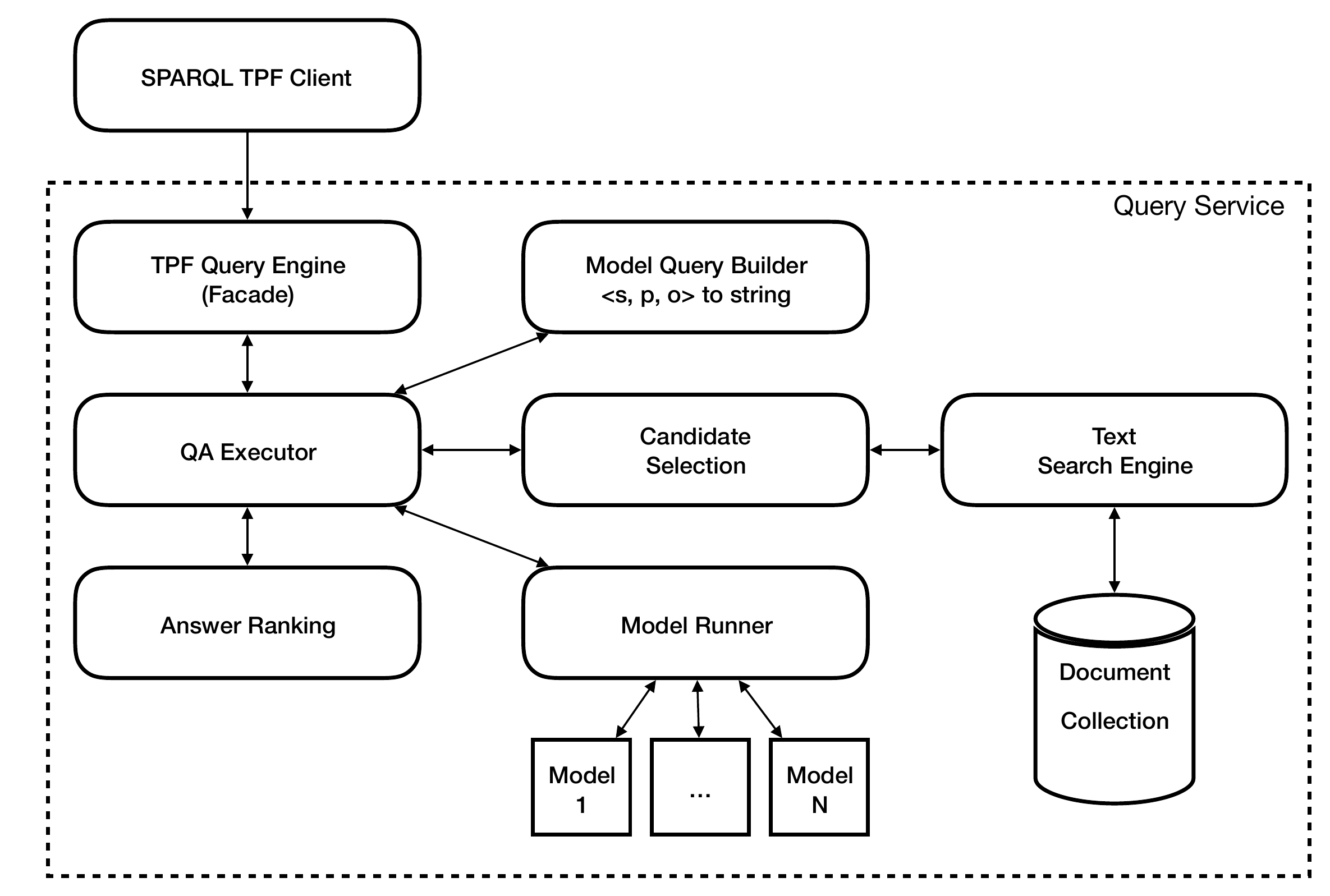}
\caption{Components of the overall system for structured query answering over text.\label{fig:systemdiagram}}
\end{figure}

Our overall approach consists of several components as illustrated in Figure \ref{fig:systemdiagram}. First, structured queries as expressed by SPARQL~\cite{harris2013sparql} are executed using a Triple Pattern Client (SPARQL TPF Client). Such a client breaks down a more complex SPARQL query into a series of triple patterns that are then issued to a service. Triple patterns are queries of the form subject, predicate, object, where each portion can be bound to an identifier (i.e. URI) or a variable.\footnote{Objects can also be bound to a literal.} Within the service, the execution engine (QA Executor) first lexicalizes the given identifiers into strings using the Model Query Builder component. For example, this component would translate an identifier like https://www.wikidata.org/wiki/Q727 into the string form ``Amsterdam''.  These queries are then issued to a candidate selection component. This component queries a standard text search engine to find potential documents that could contain the answer to the specified query. 

The candidate documents along with the lexicalized queries are provided to a model runner which issues these to models trained specifically to bind the variable that is missing. That is given a query of the form $<s, p, ?o>$ where s and o are bound and o is the variable, there would be specific models trained to extract $?o$ from the provided document. For example, given the query (\textsc{:Amsterdam} \textsc{:capital\_of} $?o$) we would have models that know how to answer queries of where the type of the subject is \textsc{City} and the property is \textsc{capital\_of}. Likewise, there would be models of that are able to answer queries of the form $<?s, p, o>$ and so on. Each model is then asked to generate a binding of the variable. Note that the bindings generated by the models are strings. The results of each model are then ranked (Answer Ranking). Using a cut-off, the results are then translated back into identifier space and returned to the client. 

A key insight of our approach is that by breaking down complex queries into triple patterns we can simplify the queries that need to be answered by the learned models. 

Our approach relies on the construction of models that are able to extract potential candidate answers from text. Following from \cite{jackthereader2018} and \cite{kumar2016ask}, we cast the problem in terms of a question answering task, where the input is a question (e.g. entity type + relation) and a document and the output is answer span within the document that binds the output. To learn these sorts of models we construct training data from knowledge graphs that have a corresponding representation in text. In the next section, we go into detail about the construction of the necessary training data. 

\section{Training Data Construction}
We begin by describing the data source employed and then describe a generic construction method in order to highlight what general data set features are needed in order to obtain the required training data. Details about the resulting training data generated using the method are then given.

\subsection{Data Sources}
Our training data is based on the combination of Wikidata and Wikipedia. Wikidata is a publicly accessible and maintained knowledge base of encyclopedic information~\cite{vrandevcic2012wikidata}. It is a graph structured knowledge base (i.e. a  knowledge graph) describing entities and the relations between them. Every entity has a globally unique identifier. Entities may also have attributes which have specific datatypes. Entities have may have more than one type. Relations between entities may hold between differing entity types.

Wikidata has a number of properties that make it useful for building a corpus to learn how to answer structured queries over text. First, and perhaps most importantly, entities have a parallel description with Wikipedia. By our count, Wikidata references 7.7 million articles in the English language Wikipedia. Thus, we have body of text which will also most likely contain answers that we retrieve from Wikidata. Second, every entity and relation in Wikidata has a human readable label in multiple languages. This enables us to build a direct connection between the database and text. Third, Wikidata is large enough to provide for adequate training data in order to build models. Finally, Wikidata provides access to their data in a number of ways including as a SPARQL endpoint, a triple patterns fragment endpoint and as a bulk RDF download. 

We did note some issues with the content of these data sources: 1) Wikidata entities are not always present in the Wikipedia content, sometimes due to using a different lexical form other times they are simply absent; 2) the Wikipedia content we used for extraction did not contain all elements of the page, notably the infoboxes were not present and our extraction pipeline did not cope with Wikipedia template markup which stripped any embedded text which are typically parameters to the template function.

Also, we chose to take static snapshots of these sources with as small a gap between them as possible. This is to minimize any differences in the sources and also to avoid issues when rerunning the processes. As one example during our development period Canada was labelled a country one day and then it was not.\footnote{After noticing this we actually modified Wikidata to correct this change.}

While we use Wikidata, we believe that our approach can be extended to any knowledge graph that has textual sources.

\subsection{Construction Method}
To describe the method, we first define our notation more formally. We adopt a view of the database as represented using simple RDF as described in the formalization given in \cite{perez2009semantics}. 

Assume there are pairwise disjoint infinite sets $I$ (IRIs),  $B$ (Blank nodes), and $L$ (Literals). An RDF term is an element in the set $T = I \cup B \cup L$. A tuple $(s, p, o) \in (I \cup B) \times I \times T$ is called an RDF triple. In this tuple $s$ is called the subject, $p$ the predicate and $o$ the object. Functional accessor notation is used to detonate the subject, predicate, or object of a triple (i.e. $t[s]$ returns the subject of a triple $t$). Also, assume a set of variables, $V$, disjoint from $T$ which are denoted by a labelling with a '?' symbol.  An RDF graph is a set of RDF triples. We refer to an RDF graph as a \textit{dataset}. 

For a query language, we adopt Triple Pattern Fragments (TPF)~\cite{tpfjournal2016}. Formally, a \textit{triple pattern} is a tuple $tp = (I \cup V) \times (I \cup V) \times (I \cup L \cup V)$. This is also called a \textit{graph pattern}. Following \cite{hartig2017formal}, TPF is a language consisting of single triple patterns. We note that triple pattern fragments are building blocks to answer much more sophisticated queries including a majority of SPARQL~\cite{tpfjournal2016}. The \textit{query result} of a triple pattern over a dataset, denoted by $[tp]_{DATASET}$, consists of a set of partial mappings $u: V \rightarrow T$. $u[tp]$ denotes a triple that is obtained by replacing variables in $tp$ according to $u$, A triple, $t$, is a \textit{matching triple} for a triple pattern if there exists a mapping $u$ such that $t = u[tp]$. 

For simplicity, we define the following functions:
\begin{itemize}
\item  $type: I \rightarrow (I \cup \emptyset)$ which returns a set of all types for a given entity with an IRI. 
\item $lexicalize: I \rightarrow (L \cup \emptyset)$ which returns the string label of a given IRI. We use this to lexicalize entity ids. 
\item $textual\_description: I \rightarrow (L \cup \emptyset)$ which returns a string containing a textual description of an entity. For example, in the case of Wikidata, this would be the contents of the Wikpedia page describing the entity. 
\item $anchor: L \times L \rightarrow ((N \times N ) \cup \emptyset ) $ which given a string returns an offset location within the other string provided.
\end{itemize}

The aim of the method is to generate  datasets of the form: [\textsc{query}; \textsc{answer}; \textsc{text in which the query is answered}]. As previously mentioned, complex queries can be expressed as a series of graph patterns. Thus, the queries we consider are graph patterns in which two of the variables are bound (e.g. \textsc{:New\_England\_Patriots} \textsc{:play} $?x$). We term these \emph{slot filling} queries as the aim is to bind one slot in the relation (i.e. the subject or the object). While we do not test graph patterns where the predicate is the variable, the same approach is also applicable.  In some sense, one can think of this as generating data that can be used to build models that act as substitute indexes of a database.

\begin{algorithm}[t]\footnotesize
\begin{algorithmic}[1]
\REQUIRE d: a dataset
\REQUIRE MAX\_TYPE\_PAIRINGS: the maximum number of paired types for a predicate that should be considered
\REQUIRE MAX\_EXAMPLES: the maximum number of examples per type pair per predicate
\STATE $results \gets Map[]$ \COMMENT{A map from a predicate IRI to a set of examples}
\FORALL {p in d}  {
	\STATE $predicate\_examples$, $subj\_types$, $obj\_types \gets \emptyset$
	\STATE $type\_pairs\_frequency \gets List[ ]$
	\STATE $triples\_per\_property \gets [(?s, p, ?o)]_d$
	\FORALL {$t \in triples\_per\_predicate$ } 
  		\STATE $subj\_types \gets subj\_types \cup type(t[s])$
		\STATE $obj\_types \gets obj\_types \cup type(t[o])$
	\ENDFOR
	\FORALL {$(st, ot) \in subj\_types \times obj\_types$} 
    		\STATE $c \gets 0$
    		\FORALL {$t \in triples\_per\_predicate$} 
			\IF {$type(t[s]) = st$ \AND $type(t[o]) = ot$} 
    				\STATE $c \gets c + 1$
    			\ENDIF
  		\ENDFOR
		\STATE append $((st,ot), c)$ to $type\_pairs\_frequency$
	\ENDFOR
	\STATE sort $type\_pairs\_frequency$ on $c$
	\FOR{$i=0$ to MAX\_TYPE\_PAIRINGS} 
		\STATE $(st,ot) \gets type\_pairs\_frequency[i][0]$
		\FOR {$j$ to MAX\_EXAMPLES}
			\FORALL {$t \in triples\_per\_property$} 
				\IF {$type(t[s]) = st$ \AND $type(t[o]) = ot$} 
        					\STATE $question \gets lexicalize(t[s])$ concatenate $lexicalize(t[p])$
					\STATE $answer \gets lexicalize(o)$
					\STATE $text \gets textual\_description(t[s])$ \label{alg:textlookup}
					\STATE $anchor \gets determine\_anchor(text, answer)$
					\STATE $example \gets \{[t, question, answer, text, anchor]\}$
					\STATE $predicate\_examples \gets predicate\_examples \cup 	example$	\label{alg:rowtd}	\ENDIF	
			\ENDFOR					
		\ENDFOR			
	\ENDFOR
	\STATE $results[p] \gets predicate\_examples$
} \ENDFOR
\RETURN results
\end{algorithmic}
\caption{Extraction method for taking a knowledge graph and generating training data.}
\label{alg:extractionmethod}
\end{algorithm}

Algorithm \ref{alg:extractionmethod} specifies the extraction method.  Given a graph dataset, the algorithm loops through all of the predicates (i.e. relations) in the dataset. It determines the frequency with which a predicate connects different types of entities. This is essential as large knowledge graphs can connect many different types using the same predicate. Thus, examples from different types of subjects and objects are needed to capture the semantics of that predicate. Using the most frequently occurring pairs of entity types for a predicate, the algorithm then retrieves as many example triples as possible where the subject and object of the triple are instances of the connected types - up to a given maximum threshold (MAX\_EXAMPLES). Thresholding is used to help control the size of the training data. 

Each triple is then used to generate a row of training data (line \ref{alg:rowtd}) for learning how to answer graph pattern queries that contain the given predicate. To connect the graph pattern queries, which are expressed using entity IRIs to the plain text over which it should be answered,  each of the components of the triple is lexicalized. In Algorithm \ref{alg:extractionmethod}, the lexicalized subject and predicate of each triple are concatenated together to form a textual query and use the lexicalized object as the answer. We then retrieve the text describing the subject (line \ref{alg:textlookup}).  We assume that the text contains some reference to the object under consideration. 

The location of that reference which we term an anchor is computed by the given anchor function. For simplicity, in our implementation, we  locate the first instance of the answer in the text. This may not always represent an instance of the answer's lexical form which is located in an expression which answers the specific question form. More complex implementations could use different heuristics or could return all possible anchor locations. 

While the algorithm defines the generation of training data for learning how to answer queries where the subject and predicate are bound, it is trivially modified for graph patterns where the subject is left unbound and the object and predicate are bound. That is to say we can build models for graph patterns of the form (s , p, ?o)  and (?s, p, o). Practically, we perform this selection at training time so two sets of training data do not need to be generated.

\subsection{Training Data}
\begin{figure}
\centering
\includegraphics[width=.7\textwidth]{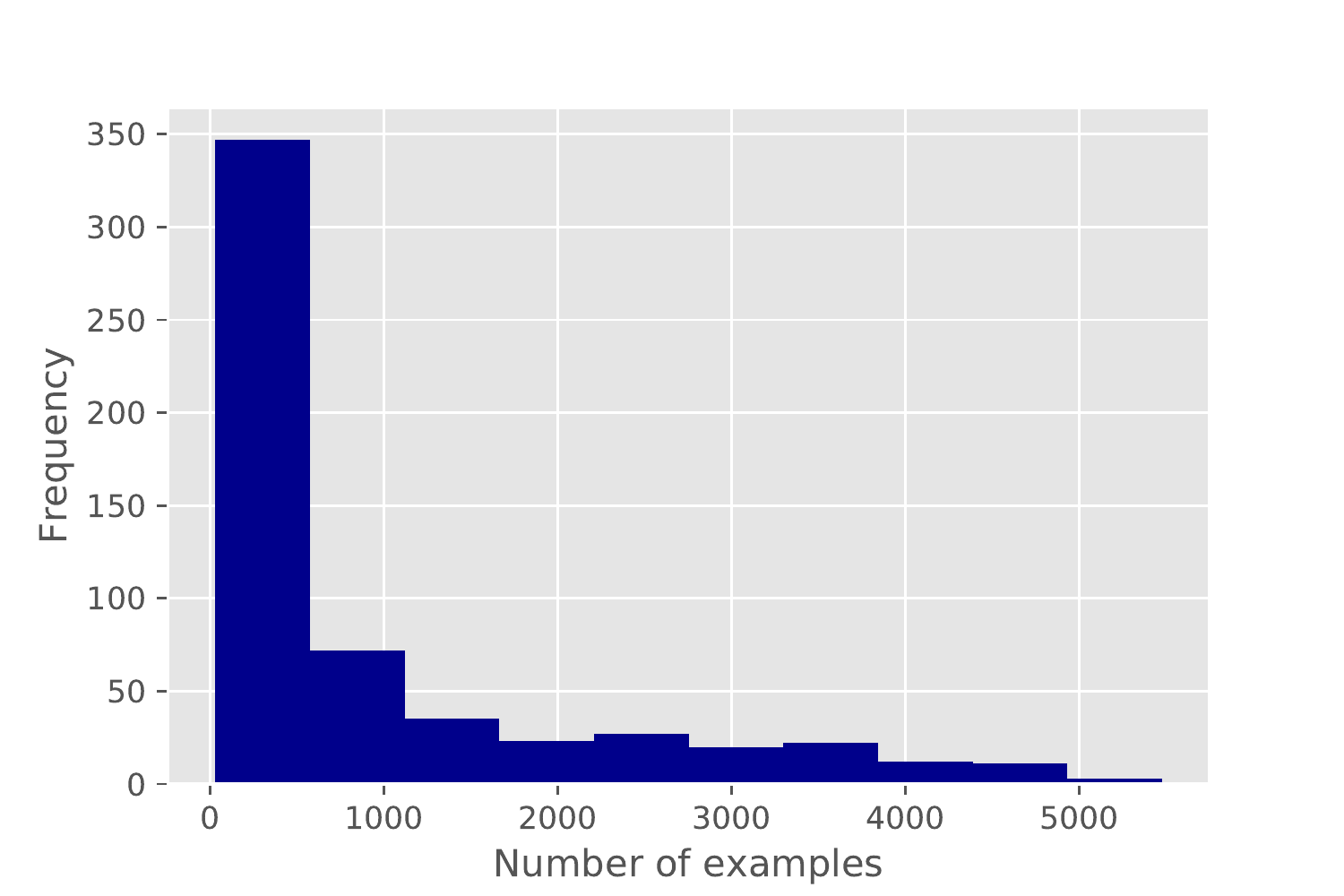}
\caption{A histogram (bin = 10) showing the grouped frequency of the amount of training data generated after cleaning. \label{fig:tdhist}}
\end{figure}

We apply the algorithm to the combination of Wikipedia and Wikidata dumps\footnote{Specifically we used Wikipedia 2018-08-20 (enwiki-20180820-pages-articles-multistream.xml.bz2) and Wikidata 2018-08-29.}. We attempted to obtain training data for all 1150 predicates in Wikidata that associate two entities together. At this time, we do not treat predicates that connect entities to literals. This is left for future work. 

Per the construction method above, we limited the extraction to the top 20 entity type pairs per predicate (MAX\_TYPE\_PAIRINGS), and limited each type pair to 300 examples (MAX\_EXAMPLES). Thus, there is a maximum yield of 6000 examples per predicate. We then apply the following cleaning/validation to the retrieved examples. 

First, we drop examples where there is no Wikipedia page. Second, we ensure that the answer is present in the Wikipedia page text. Finally, in order to ensure adequate training data we filter out all models with less than 30 examples.  Note that this means that we have differing amounts of training data per predicate.

After cleaning, we are able to obtain training data for 572 predicate for the setting in which the object is the variable/answer. We term this the SP setting. On average we have 929 examples per predicate with a maximum number of examples of 5477 and a minimum of 30 examples. The median number of examples is 312. Figure \ref{fig:tdhist} shows the frequency of training data across the predicates. 

In the setting in which the subject is the variable / answer we are trying to extract, enough data for 717 predicates is obtained. This is because the subject answer is more likely to appear in the Wikipedia page text. We term this the PO setting.

\section{Models}
Based on the above training data, we individual train models for all predicates using the Jack the Reader framework~\cite{jackthereader2018}. We use two state-of-the-art deep learning architectures for extractive question answering, namely, FastQA~\cite{weissenborn2017making}  and the implementation provided by the framework, JackQA.  Both architectures are interesting in that while they perform well on reading comprehension tasks (e.g. SQuAD~\cite{rajpurkar2016squad}) both architectures try to eliminate complex additional layers and thus have the potential for being modified in the future to suit this task. Instead of describing the architectures in detail here, we refer the reader to corresponding papers cited above. We also note that the Jack the Reader configuration files provide succinct descriptions of the architectures, which are useful for understanding their construction. 

To improve performance both in terms of reducing training time and to reduce the amount of additional text the model training has to cope with, we applied a windowing scheme. This is because longer text is normally associated with greater issues when dealing with sequence models. Our scheme takes a portion of the text around the answer location chosen from the Wikipedia content.

We now describe the following parameters for each architecture. 

\paragraph{FastQA} All text is embedded using pre-trained GloVe word embeddings~\cite{pennington2014glove} (6 billion tokens, and 50 dimensions). We train for 10 epochs using a batch size of 20.  We constrain answers to be a maximum of 10 tokens and use a window size of 1000 characters. The answer layer is configured to be bilinear. We use the ADAM optimizer with a learning rate of 0.11 and decay of 1.0. 

\paragraph{JackQA} Here we embed the text using pre-trained GloVe word embeddings (840 billion tokens and 300 dimensions). We use the default JackQA settings. We use a window size of 3000 characters. The batch sizes were 128/96/64 for three iterative runs. The subsequent runs with smaller batch sizes were only run if the prior iteration failed. We specified a maximum number of 20 epochs. 

\paragraph{Baseline}
In addition to the models based on neural networks, we also implemented a baseline. The baseline consisted of finding the closest noun phrase to the  property within the Wikipedia page and checking whether the answer is contained within that noun phrase. 

Note, we attempted to find functional settings that worked within our available computational constraints. For example, FastQA requires more resources than JackQA in relation to batch size , thus, we chose to use smaller embeddings and window size in order to maintain a ``good" batch size. 

We use 2/3 of the training data for model building and 1/3 for testing. Data is divided randomly. Training was performed using an Amazon EC2 p2.xlarge\footnote{1 virtual GPU - NVIDIA K80, 4 virtual CPUs, 61 GiB RAM} box. It took ~23 hours for training of FastQA models, which included all models for all predicates even when there were too few training samples. For JackQA, the window was increased to 3000 characters, and multiple training sessions were required, reducing the batch size each time to complete the models which  not finish from earlier runs, in all three passes were required with 128, 96 and 64 batch size respectively. Total training time was ~81 hours.

Note that we train models for the setting where the subject and predicate are bound but the object is not. We also use the FastQA architecture to build models for the setting where the subject is treated as the variable to be bound. 

%
%
%

\section{Experimental Results}
\begin{table}
\centering
\begin{tabular}{lcrrrrrrr}
Model &  Model Count &  mean &   std &  min &  max &   25\% &   50\% &   75\% \\
\hline
JackQA - SP   &  572 &  0.70 &  0.24 &  0.0 &  1.0 &  0.54 &  0.77 &  0.89 \\
FastQA - SP   &  572 &  0.62 &  0.24 &  0.0 &  1.0 &  0.43 &  0.65 &  0.80 \\
FastQA - PO &  717 &  0.89 &  0.10 &  0.4 &  1.0 &  0.85 &  0.92 &  0.96 \\
Baseline &  407 & 0.15 & 0.17 & 0.0 & 0.86 & 0.03 & 0.08 & 0.20 \\
\hline
\end{tabular}
\caption{F1 results across all models and the baseline \label{table:f1results}}
\end {table}

\begin{table}
\centering
\begin{tabular}{lcrrrrrrr}
Model &  Model Count &  mean &   std &  min &  max &   25\% &   50\% &   75\% \\
\hline
JackQA - SP    &  572 &  0.64 &  0.26 &  0.0 &  1.0 &  0.44 &  0.71 &  0.86 \\
FastQA - SP   &  572 &  0.55 &  0.25 &  0.0 &  1.0 &  0.36 &  0.57 &  0.74 \\
FastQA - PO &  717 &  0.83 &  0.14 &  0.1 &  1.0 &  0.75 &  0.86 &  0.94 \\
\hline
\end{tabular}
\caption{Exact results \label{table:exactresults}}
\end {table}

Table \ref{table:f1results} one reports the average F1 measure across all models as well as the baseline. This measure takes into account the overlap of the identified set of tokens with the gold standard answer controlling for the length of the extracted token. By definition, the baseline only generates such overlap scores.  

Table \ref{table:exactresults} reports the average exact match score over all models. This score measures whether the model extracts the exact same string as in the gold standard. For reference, both tables also reports the total number of models trained (Model Count), which is equivalent to the training data provided. The Model Count for the baseline is equivalent to the number of predicates for which the baseline method could find an answer for.


\begin{figure}
\centering
\begin{subfigure}[b]{0.49\textwidth}
\includegraphics[width=\textwidth]{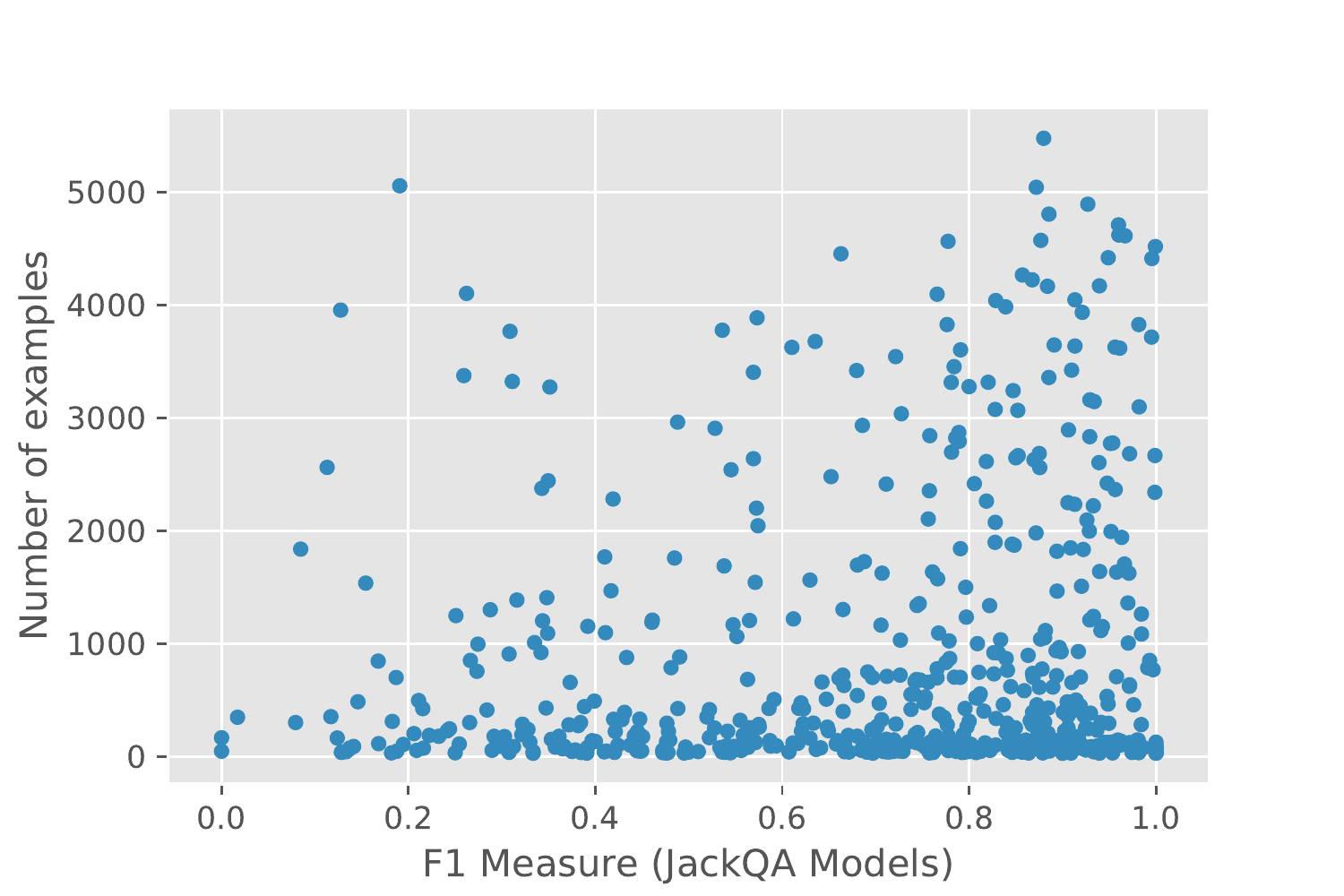}
\caption{Models trained using Jack QA}
\end{subfigure}
\begin{subfigure}[b]{0.49\textwidth}
\includegraphics[width=\textwidth]{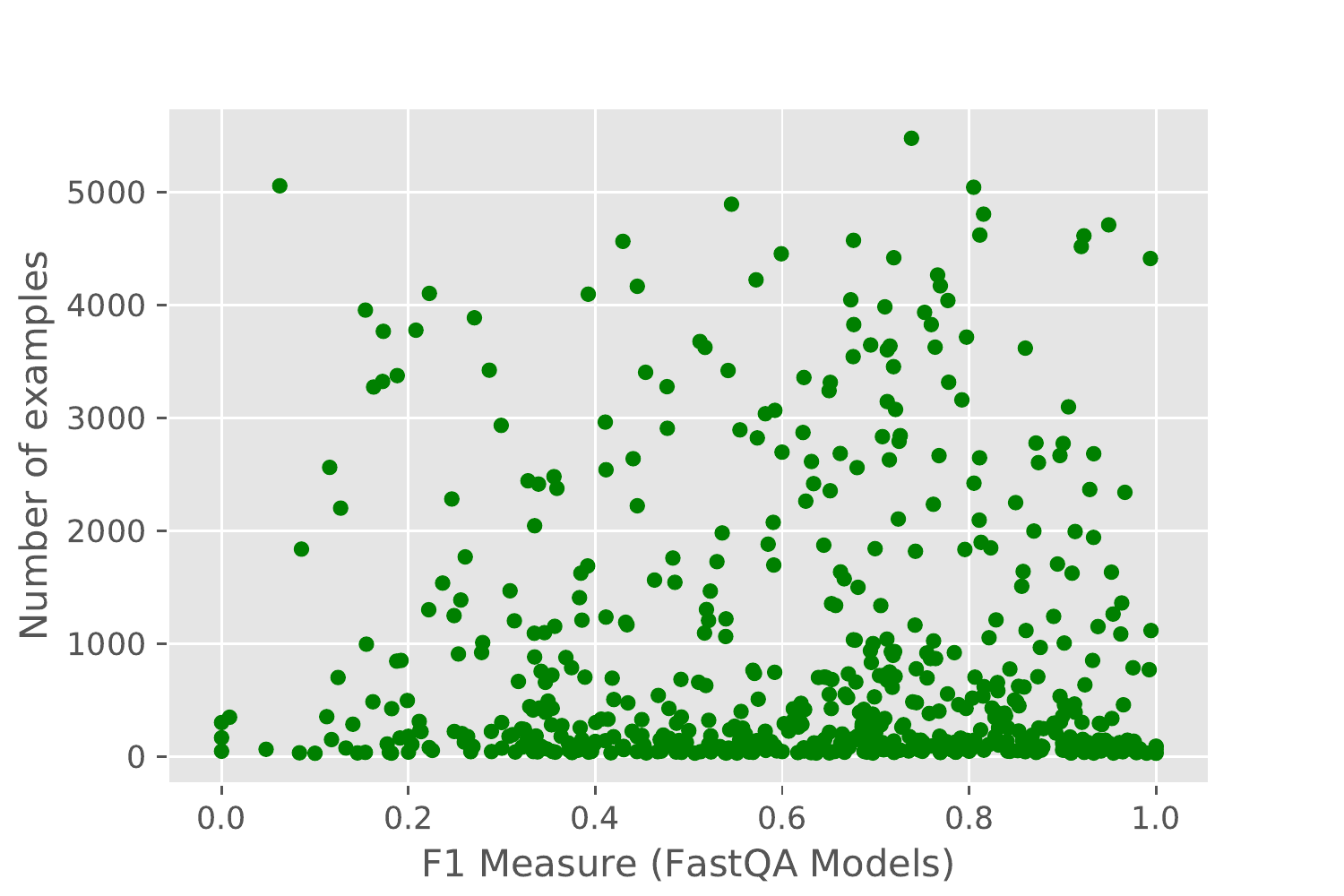}
\caption{Models trained using Fast QA.}
\end{subfigure}
\caption{Plot of individual model performance vs. training data size. All 572 models are shown for the SP setting.\label{fig:tdperform}}
\end{figure}

Figure \ref{fig:tdperform} plots individual model performance against the size of the training data given. Overall, models based on deep learning notably outperform the baseline models on average. Additionally, using these deep learning based approaches we are able to create models that answer queries for 160 additional properties over the baseline. 

\section{Analysis}
First, we wanted to see if there was a correlation between the amount of training data and the performance of a model. Using the data presented in Figure \ref{fig:tdperform}, we fit a linear regression to it. We found no statistically significant correlation ($R^2  = 0.37$).

\begin{figure}
\centering
\begin{subfigure}[b]{0.49\textwidth}
\includegraphics[width=\textwidth]{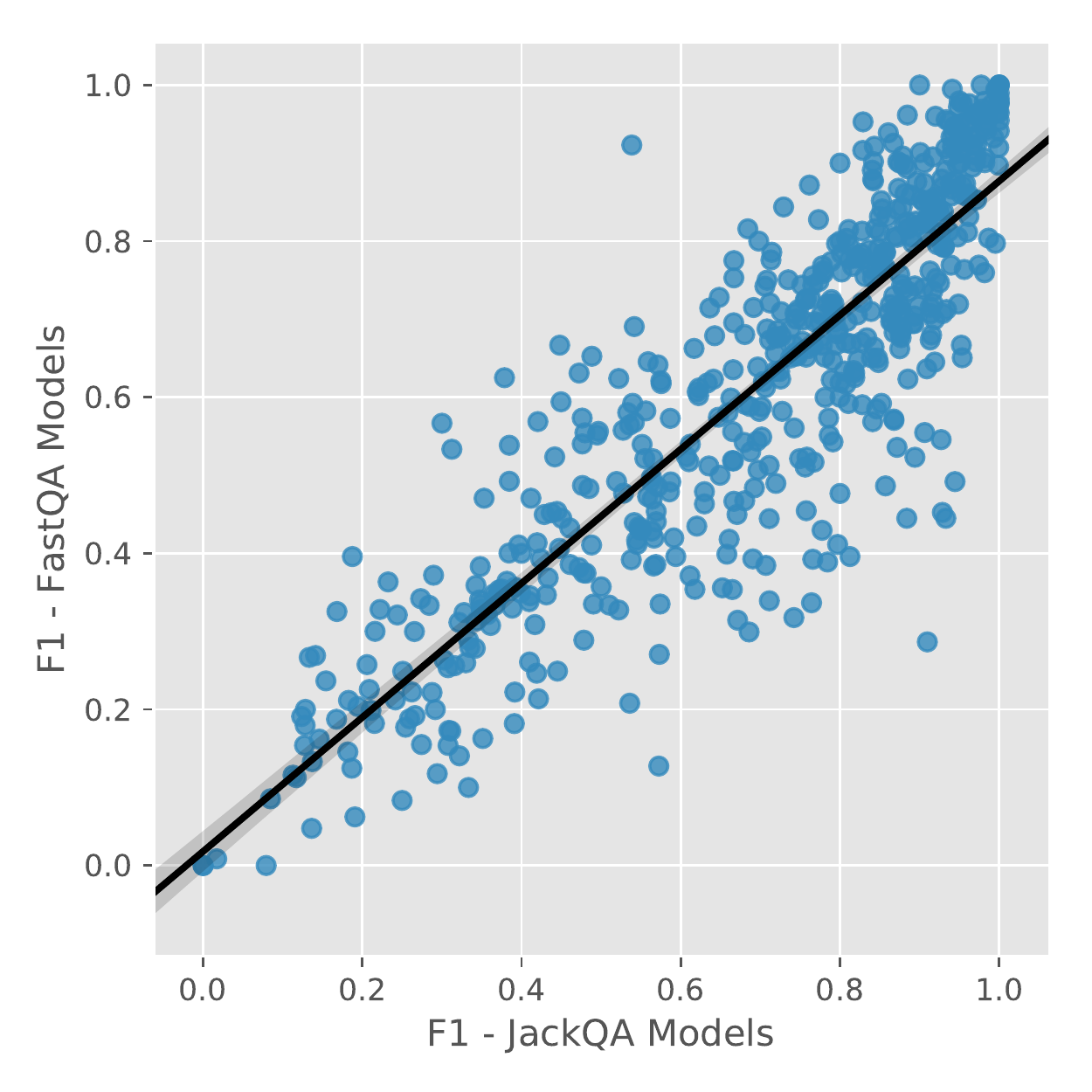}
\caption{Comparison of F1 score performance \label{fig:f1correlation}}
\end{subfigure}
\begin{subfigure}[b]{0.49\textwidth}
\includegraphics[width=\textwidth]{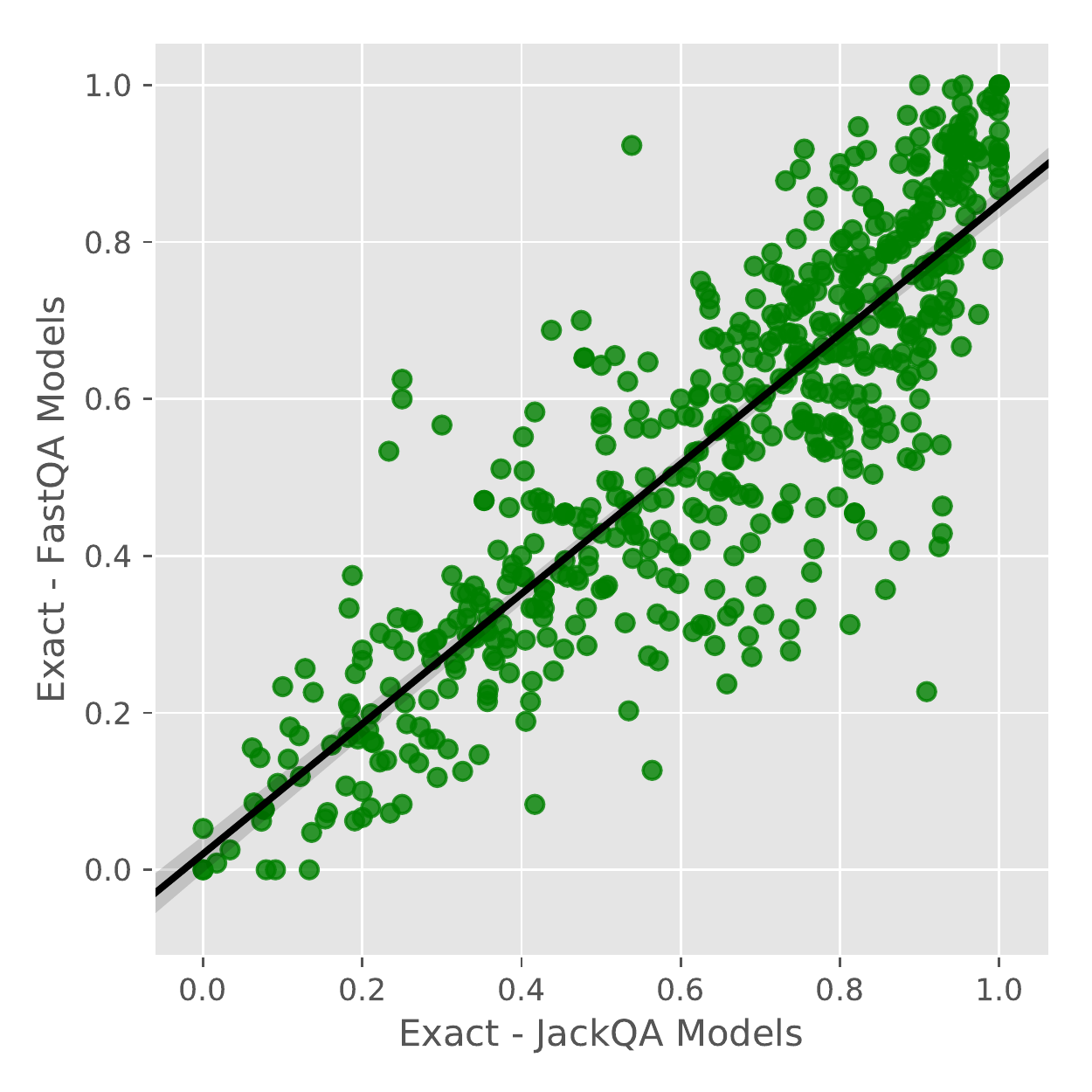}
\caption{Comparison of Exact score performance \label{fig:exactcorrelation}}
\end{subfigure}
\caption{Comparison of the performance on all properties using different model architectures in the SP setting. \label{fig:compscores}}
\end{figure}

The model architectures show strong correlation in performance (c.f. Figure \ref{fig:compscores}). The $R^2$ value being $0.97$ in the case of the F1 measure and $0.96$ for the Exact measure. This suggests that the performance is primarily a factor of the underlying kind of data. 

Therefore, we looked more deeply at performance for individual models for a given property. Table \ref{table:highperf} shows the highest performing models. We find some consistent patterns. First, properties that have specific value constraints within Wikidata generate good results. For example, the ``crystal system'' property needs to have one of 10 values (e.g cubic crystal system, quasicrystal, amorphous solid). Likewise, the ``coolant'' property needs to be assigned one of fourteen different values (e.g. water, oil, air).  This is also true of ``discovery method'', which oddly enough is actually defined as the the method by which an exoplanet is discovered. This is also a feature of properties whose values come from classification systems (e.g. ``Köppen climate classification'' and ''military casualty classification''). 

A second feature that seems to generate high performing models are those that refer to common simple words. For example, the ``source of energy'' property takes values such as ``wind''  or ``human energy''. 

Lastly, simple syntactic patterns seem to be learned well. For example, the property ''birthday'', which links to entities describing a month, day combination (e.g. November 8) which is thus restricted to a something that looks like a month string followed by one or two numerical characters. Likewise, the expected value for the property ``flag'' often appears directly in text itself. That is the correct answer for the query ``Japan flag'' is ``flag of Japan'', which will appear directly in text. 

\begin{table}
\centering
\begin{tabular}{p{4.3cm}rrrrr}
Property &  Fast QA &  Fast QA &  Jack QA &  Jack QA & Training \\
& F1 & Exact & F1 & Exact & Data Size \\     
\hline
birthday &    0.95 &       0.91 &     1.0 &        1.0 &          32 \\
 flag &    0.98 &       0.88 &     1.0 &        1.0 &          50 \\
league points system &    1.00 &       1.00 &     1.0 &        1.0 &          90 \\
discovery method &    0.98 &       0.91 &     1.0 &        1.0 &          69 \\
source of energy &    0.94 &       0.94 &     1.0 &        1.0 &          50 \\
military casualty classification &    1.00 &       1.00 &     1.0 &        1.0 &          92 \\
topic's main category &    0.99 &       0.91 &     1.0 &        1.0 &          31 \\
Köppen climate classification &    1.00 &       1.00 &     1.0 &        1.0 &          34 \\
coolant &    0.98 &       0.98 &     1.0 &        1.0 &         128 \\
crystal system &    0.96 &       0.87 &     1.0 &        1.0 &          43 \\
\hline
\end{tabular}
\caption{Highest 10 performing models in the SP setting as determined by F1 measures from models trained using the Jack QA architecture. \label{table:highperf}}
\end{table}

We also look at the lowest performing models, shown in Table \ref{table:lowperf} to see what is difficult to learn. Ratings for films (e.g.  Australian Classification, RTC film rating, EIRIN film rating) seem extremely difficult to learn. Each of these properties expect values of two or three letters (e.g. PG, R15+, M). The property ``blood type'' also has the same form.  It seem that using character level embeddings may worked better in these cases. 

The property ``contains administrative territorial entity '' is an interesting case as there are numerous examples. This property is used within Wikidata to express the containment relation in geography. For example, that county contains a village or a country contains a city. We conjecture that this might be difficult to learn because the sheer variety of linkages that this can express making it difficult to find consistencies in the space.  A similar issue could be present for properties such as ``voice actor'' and ``cast member'' where the values can be essentially any person entity. Similarly, ``polymer of'' and ``species kept'' both can take values that come from very large sets (e.g. all chemical compounds and all species). It might be useful for the model to be provided specific hints about types (i.e. actors, chemicals, locations) that may allow it to find indicative features. 

\begin{table}
\centering
\begin{tabular}{p{4.3cm}rrrrr}
Property &  Fast QA &  Fast QA &  Jack QA &  Jack QA & Training \\
& F1 & Exact & F1 & Exact & Data Size \\
\hline
Australian Classification &    0.00 &       0.00 &    0.00 &       0.00 &          48 \\
RTC film rating &    0.00 &       0.00 &    0.00 &       0.00 &         167 \\
EIRIN film rating &    0.01 &       0.01 &    0.02 &       0.02 &         349 \\
blood type &    0.00 &       0.00 &    0.08 &       0.08 &         302 \\
contains administrative territorial entity &    0.09 &       0.06 &    0.08 &       0.07 &        1838 \\
voice actor &    0.12 &       0.11 &    0.11 &       0.09 &        2562 \\
species kept &    0.11 &       0.03 &    0.12 &       0.03 &         354 \\
best sprinter classification &    0.19 &       0.18 &    0.12 &       0.11 &         165 \\
cast member &    0.15 &       0.14 &    0.13 &       0.11 &        3955 \\
 polymer of &    0.18 &       0.08 &    0.13 &       0.08 &          38 \\
\hline
\end{tabular}
\caption{Lowest 10 performing models in the SP setting as determined by F1 measures from models trained using the Jack QA architecture. \label{table:lowperf}}
\end{table}

\section{Prototype}
To understand whether this approach is feasible in practice,  we implemented a prototype of the system outlined in Figure \ref{fig:systemdiagram}. For the triple pattern fragment facade we modify Piccolo, an open source triple pattern fragments server to replace its in-memory based system with functions for calling out to our QA answering component. The facade also implements a simple lexicalization routine. The query answering component is implemented as a Python service and calls out to an Elasticsearch\footnote{https://github.com/elastic/elasticsearch} search index where documents are stored. The query answering component also pre-loads the models and runs each model across candidate documents retrieved by querying elastic search. We also specify a max number of candidate documents to run the models over. Currently, we execute each model sequentially over all candidate documents. We then chose the top set of ranked answers given the score produced by the model. Note that we can return multiple bindings for the same ranked results. 

We measured the performance of our system over Wikipedia. It takes on the order of 10 seconds to provide results for a single triple pattern query. This is surprisingly good given the fact that we execute models sequentially instead of in parallel. Furthermore, we execute the models over the entirety of the Wikipedia article. Our own anecdotal experience shows that question answering models are both faster and produce more accurate results when supplied with smaller amounts of text. Thus, there is significant room for optimizing query performance with some simple approaches including parallelizing models, chunking text into smaller blocks, and limiting the number of models executed to those that are specific for the triple pattern. Furthermore, it is straightforward to issue triple pattern fragment query requests over multiple running instances~\cite{tpfjournal2016}. One could also implement more complex sharding mechanisms designed for triple tables~\cite{abdelaziz2017survey}. Overall, the prototype gives us confidence that this sort of system could be implemented practically.\footnote{We also integrated the prototype with Slack.} 

\section{Related Work}
Our work builds upon and connects to a number of existing bodies of literature. The work on information extraction is closely related. \cite{martinez2018information} provides a recent survey of the literature in this area specifically targeted to the the problems of extracting and linking of entities, concepts and relations. One can view the models that we build as similar to distantly supervised relation extraction approaches~\cite{mintz2009distant, surdeanu2012multi}, where two mentions of entities are found in text and the context around those mentions is used to learn evidence for that relation. Recent approaches have extended the notion of context~\cite{quirk2017distant} and applied neural networks to extract relations \cite{zeng2014relation, glass2018inducing}.  

The closest work to ours  in the information extraction space is \cite{levy2017zero} where they apply machine comprehension techniques to extract relations. Specifically, they translate relations into templated questions - a process they term querification. For example, for the relation spouse(x,y) they created a series of corresponding question templates such as ``Who is x married to?''. These templates are constructed using crowdsourcing, where the workers are provided a relation, example sentence and asked to produce a question template. This dataset is used to train a BiDAF-based model \cite{seo2016bidirectional} and similar to our approach they address slot filling queries where the aim is to populate one side of the relation. The model achieves good results resulting in an 89\% F1 measure when predicting relations between entities. While we apply a similar technique,  our approach differs in a number of aspects. First, we target a different task, namely, answering structured queries. Second, we do not generate questions through question templates but instead build the questions out of the knowledge base itself. Third, instead of training a joint model for relations, we train a unique model for every relation. This approach fits well to our task which is aiming to bind triple patterns. The training of individual models has been in effective in other tasks~\cite{hoffmann2015extreme}. However, we believe a joint model is worth exploring for our task. 

Like much of the work in this space our approach is based on a large scale parallel corpus. Of particular relevance to our task are the WikiSQL and WikiReading corpora. WikiSQL~\cite{zhongSeq2SQL2017} provides a parallel corpus that binds SQL queries to a natural language representation. The task the dataset is used for is to answer natural language questions over SQL unlike ours which is to answer SQL-like queries over text. SQLWikiReading~\cite{hewlett2016wikireading} like our approach extracts a corpus from Wikidata and Wikipedia in order to predict the value of particular properties. Another corpus of note is ComplexWebQuestions~\cite{talmor2018web}, which pairs complex SPARQL queries with natural language queries. Importantly, it looks at the compositionality of queries from smaller units. Like WikiSQL, it looks at answering natural language queries over databases. In general, we think our approach in also specifying an extraction procedure is a helpful addition for applying corpus construction in different domains. 

As mentioned in the introduction, text databases, where information extraction is combined with databases are also relevant. Our system architecture was inspired by the pioneering work of \cite{jain2007sql}. In that work, a search index is used to first locate potential documents and then information extraction techniques are applied to the selected documents to populate a database. Our approach differs in two key aspects. First, instead of populating a database our system substitutes the indexes of the database with models. Second, we use distributed query techniques in order to process complex queries on the client side. Recent work \cite{kilias2018idel} uses deep learning based approaches to perform information extraction during database query execution specifically for entity disambiguation. Similar to other work in this area, and unlike ours, they integrate the information extraction within the database engine itself.

Finally, there is a long history of mixing information retrieval and database style queries together. For example, for the purposes of querying over semistructured data \cite{abiteboul1997querying}. \cite{raghavan2001integrating} provides an accessible introduction to that history. While our system is designed to answer database queries one can imagine easily extending to the semistructured setting. 

\section{Conclusion \& Future Work}
In this work, we have explored the notion of answering database queries over text absent the need for a traditional database intermediary. We have shown that this approach is feasible in practice by combining machine comprehension based models with distributed query techniques.

There are a number of avenues for future work. In the short term, the developed models could be expanded to include extracting properties as well as subjects and objects. We also think that joint models for all triple pattern predictions is worth exploring. One would also want to extend the supported queries to consider not only relationships between entities but also to the attributes of entities. Our current lexicalization approach is also quite simple and could be improved by considering it as the inverse of the entity linking problem and applying those techniques or applying summarization approaches~\cite{VOUGIOUKLIS2018}.  In this work, we used model architectures that are designed for answering verbalized questions and not database queries. Modifying these architectures may also be a direction to obtain even better performance. Obviously more extensive experimental evaluations would be of interest, in particular, extending the approach to other knowledge bases and looking more deeply at query result quality. 

In the long term, the ability to query over all types of data whether images, structured data or text has proven useful for knowledge bases~\cite{wu2018fonduer}. Extending our concept to deal with these other datatypes could be powerful -making it easy to perform structured queries over unstructured data while minimizing information extraction overhead. Additionally, because our approach is based on machine comprehension one could imagine that structured queries could be expressed using different schema relations than those present in the training data and potentially even support multiple query syntaxes.  

In general, we believe that structured queries will continue to be a useful mechanism for data professionals to both work with data and integrate information into existing data pipelines. Hence, focusing on automated knowledge base construction from the query vantage point is an important perspective to investigate. 

\bibliography{tpf}
\bibliographystyle{plainnat}

\end{document}